\pdfoutput=1

\documentclass[11pt]{article}

\usepackage{float}
\usepackage{placeins}
\usepackage[]{naacl2021}
\usepackage{graphicx}
\usepackage{subcaption}
\usepackage{amsmath}
\usepackage{array}
\usepackage{amsmath, nccmath}
\usepackage{geometry}
\usepackage{caption}
\usepackage{subcaption}
\usepackage{mathtools}
\usepackage{times}
\usepackage{latexsym}
\usepackage{enumerate}

\usepackage[T1]{fontenc}

\usepackage[utf8]{inputenc}

\usepackage{microtype}
\newcommand{\centered}[1]{\begin{tabular}{l} #1 \end{tabular}}
\title{Accounting for Agreement Phenomena in \\ Sentence Comprehension with Transformer Language Models:\\ Effects of Similarity-based Interference on Surprisal and Attention}

\author{Soo Hyun Ryu \\
  Department of Psychology \\
  University of Michigan \\
  \texttt{soohyunr@umich.edu} \\\And
  Richard L. Lewis\\
  Department of Psychology\\
  University of Michigan\\
  \texttt{rickl@umich.edu} \\}

\begin{document}
\maketitle
\begin{abstract}
We advance a novel explanation of similarity-based interference effects in  subject-verb and reflexive pronoun agreement processing, grounded in surprisal values computed from a pretrained large-scale Transformer  model, GPT-2. Specifically, we show that surprisal of the verb or reflexive pronoun predicts \emph{facilitatory interference effects} in ungrammatical sentences, where a distractor noun that matches in number with the verb or pronoun leads to faster reading times, despite the distractor not participating in the agreement relation. We review the  human empirical evidence for such effects, including recent meta-analyses and large-scale studies.
We also show that \emph{attention patterns} (indexed by entropy and other measures) in the  Transformer show patterns of diffuse attention in the presence of similar distractors, consistent with cue-based retrieval models of parsing. But in contrast to these models, the attentional cues and memory representations are learned entirely from the simple self-supervised task  of predicting the next word.

\end{abstract}

\section{Introduction}

Deep Neural Network (DNN) language models \cite{lecun2015deep,sundermeyer2012lstm,vaswani2017attention} have recently attracted the attention of researchers interested in assessing their linguistic competence \cite{chaves2020don,da2020assessing,ettinger2020bert,wilcox2018rnn,wilcox2019syntactic} and potential to provide accounts of psycholinguistic phenomena in sentence processing \cite{futrell2018rnns,linzen2021syntactic,van2018modeling, wilcox2020predictive}. In this paper we show how attention-based transformer models (we use a pre-trained version of GPT-2) provide the basis for a new theoretical account of facilitatory interference effects in subject-verb and reflexive agreement processing.  These effects, which we review in detail below, have played an important role in psycholinguistic theory because they show that properties of noun phrases that are not the grammatical targets of agreement relations may nonetheless exert an influence on processing time at points where those agreement relations are computed.

The explanation we propose here is a novel one grounded in surprisal \cite{hale2001probabilistic,levy2008expectation}, but with origins in graded attention and similarity-based interference \cite{van2003distinguishing,lewis2006computational,jager2017similarity}.  We use surprisal as the key predictor of reading time \cite{levy2013memory}, and through targeted analyses of patterns of attention in the transformer, show that the model behaves in ways consistent with cue-based retrieval theories of sentence processing. The account thus provides a new integration of surprisal and similarity-based interference theories of sentence processing, adding to a growing literature of work integrating noisy memory and surprisal \cite{futrell2020lossy}.  In this case, the noisy representations arise from training the transformer, and interference must exert its influence on reading times through a \emph{surprisal bottleneck} \cite{levy2008expectation}.

The remainder of this paper is organized as follows.  We first provide an
overview of some of key empirical work in human sentence processing
concerning subject-verb and reflexive pronoun agreement.
We then provide a brief overview of the GPT-2
architecture, its interesting psycholinguistic properties, and the method and metrics that we will use to examine the agreement effects.
We then apply GPT-2  to the
materials used in several  different human reading time studies.
We conclude with some theoretical reflections, identification of weaknesses, and suggestions for future work.

\section{Agreement interference effects in human sentence processing}
\begin{figure}[tb!]
\begin{center}
\includegraphics[width = 8cm, height = 6.4cm]{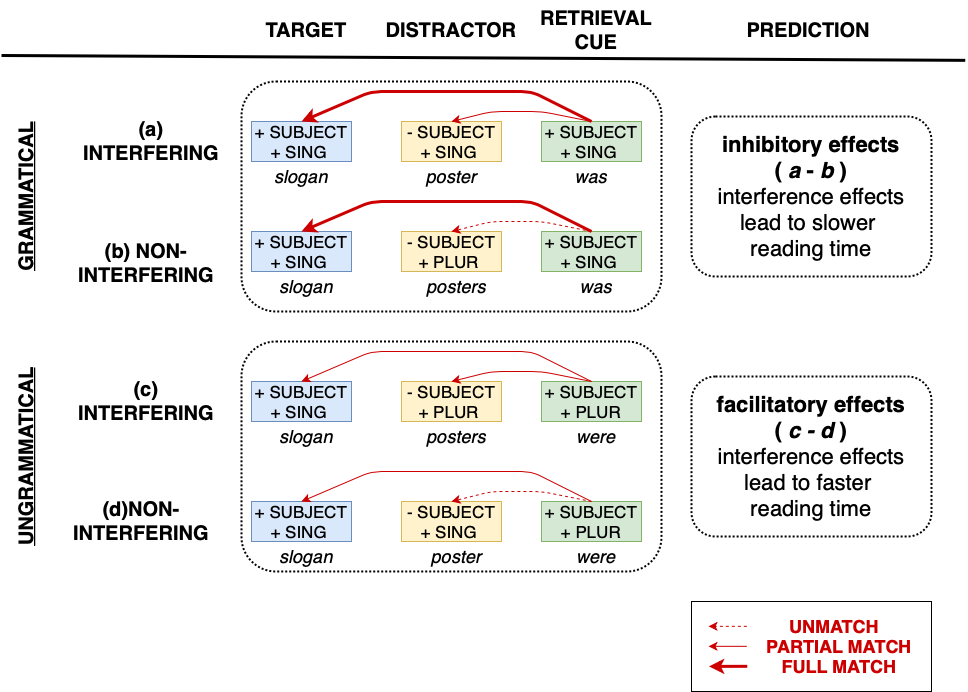}
\end{center}
\caption{How facilitatory and inhibitory interference effects arise in subject-verb dependency creation in cue-based retrieval parsing. The critical manipulation concerns the overlap of number feature between the distractor, target, and retrieval cue.}
\label{fig:scheme}
\end{figure}

One long-standing focus of work in sentence comprehension is understanding how the structure of human short-term memory might support and constrain the incremental formation of linguistic dependencies among phrases and words \cite{gibson1998linguistic,lewis1996interference,lewis2006computational,miller1963finitary, nicenboim2015working}.  A key property of human memory thought to shape sentence processing is \emph{similarity-based interference} \cite{miller1963finitary,lewis1993architecturally,lewis1996interference}. Figure \ref{fig:scheme} shows a simple example of how such interference arises in cue-based retrieval models of sentence processing, as a function of the compatibility of \emph{retrieval targets} and \emph{distractors} with retrieval \emph{cues} \cite{lewis2005activation,lewis2006computational,van2003distinguishing} (Corresponding sentences are from \citet{wagers2009agreement}'s Exp 4--6 shown in Table \ref{tab:example_set}).  \textit{Inhibitory interference effects} occur when features of the target perfectly match the retrieval cue and features of a distractor partially matches, while \textit{facilitatory interference effects} occur when the features of both target and distractor partially match the features of retrieval cue.

In this study, we focus  on interference effects in subject-verb number agreement and reflexive pronoun-antecedent agreement, specifically in languages where the agreement features include \emph{syntactic number} which is morphologically marked on the verb or pronoun. In such cases, number is plausibly a useful retrieval cue, and it is easy to manipulate the number of distractor noun phrases to allow for carefully controlled empirical contrasts.

\paragraph{Interference in subject-verb agreement.}

Previous studies \cite{pearlmutter1999agreement,wagers2009agreement,dillon2013contrasting,lago2015agreement,jager2020interference} attest to both \textit{inhibitory} interference (slower processing in the presence of an interfering distractor) and \textit{facilitatory} interference (faster processing in the presence of an interfering distractor), but the existing empirical support for inhibitory interference is weak, and many studies fail to find any evidence for it \cite{dillon2013contrasting,lago2015agreement,wagers2009agreement}. There is stronger evidence for facilitatory effects, which arise in {ungrammatical} structures where the verb or pronoun fails to agree in number with the structurally correct target noun phrase, but where either an intervening or preceding distractor noun phrase does match in number. Example A. below illustrates, taken from \citet{wagers2009agreement}, where the subject and verb are boldfaced and the distractor noun is underlined:

\begin{enumerate}[A.]
\item The \textbf{slogan} on the \underline{posters} \textbf{were} designed to get attention. 
\end{enumerate}




A Bayesian meta-analysis of agreement phenomena was recently conducted with an extensive set of studies \cite{jager2017similarity, vasishth0AD}. Their analysis of first-pass reading times from eye-tracking experiments on subject-verb number agreement is shown in Figure 1. The evidence from the meta-analysis is consistent with a very small or nonexistent inhibitory interference effect in in the grammatical conditions, with a small but robust facilitatory interference effects in the ungrammatical conditions. Concerned that the existing experiments did not have sufficient power to detect the inhibitory effects, \citet{nicenboim2018exploratory} ran a large scale eye-tracking study (185 participants) with materials designed to increase the inhibition effect, and did detect a 9ms effect (95\% credible posterior interval 0--18ms).  This represents the strongest evidence to date for inhibitory effects in grammatical agreement structures, but even this evidence indicates the effect may be near zero.

\begin{figure*}
\includegraphics[scale=.48]{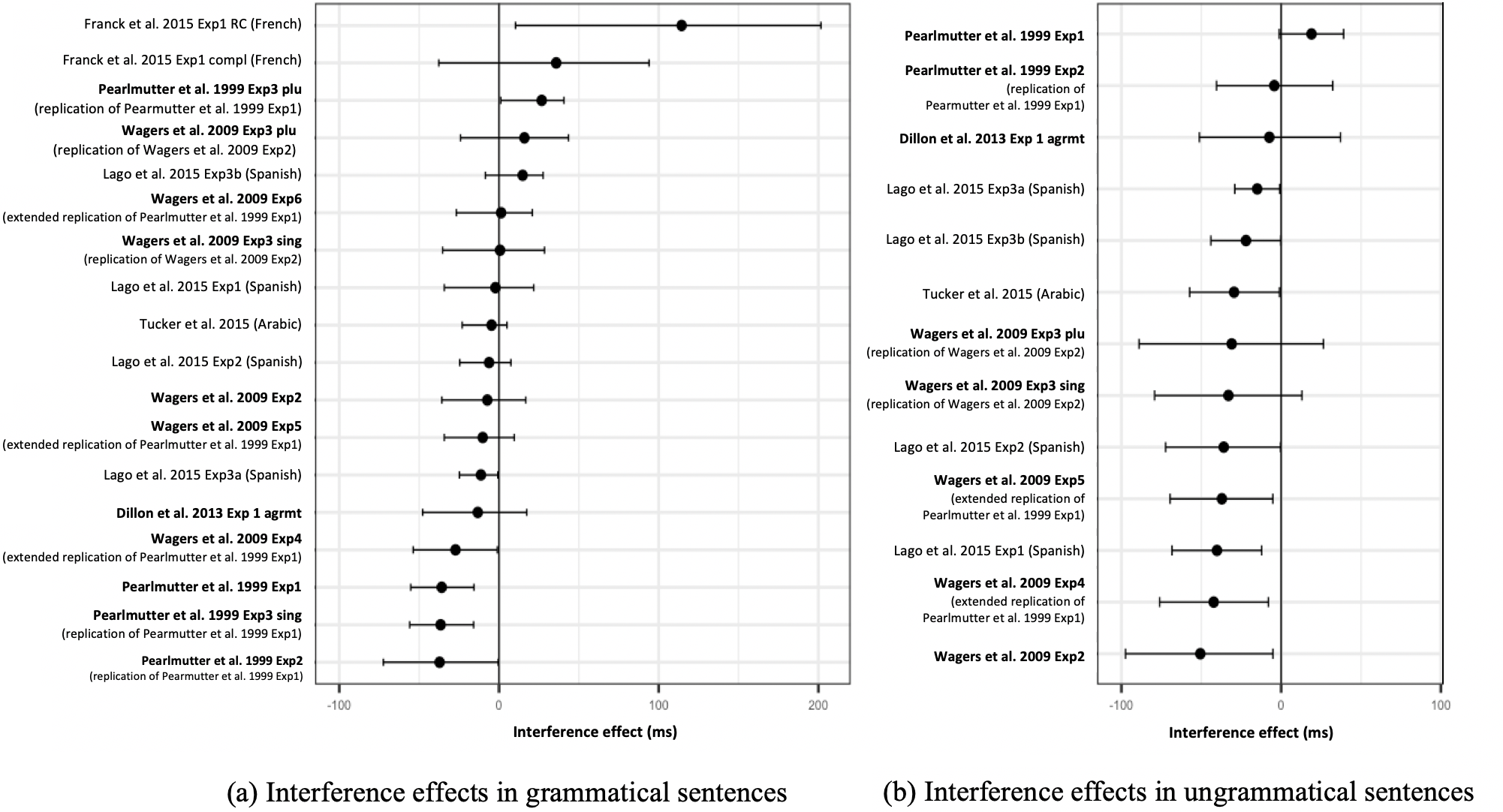}
\label{fig:subj_verb}
\hfill
\caption{Results of the meta-analysis on subject-verb number agreement from \citet{vasishth0AD}. The materials from boldfaced studies are those that we used in our GPT-2 experiments.}
\label{fig:meta_subverb}
\end{figure*}

\paragraph{Interference in reflexive pronoun agreement.}
Example B. below shows a pair of sentences from \citet{dillon2013contrasting} used to probe facilitatory effects in reflexive pronoun agreement (again, the target antecedent and pronoun are boldfaced and the distractor is underlined):

\begin{enumerate}[B.]
\item (1) \emph{interfering} The basketball \textbf{coach} who trained the star \underline{players} usually blamed \textbf{themselves} for the ...

(2) \emph{non-interfering} The basketball \textbf{coach} who trained the star \underline{player} usually blamed \textbf{themselves} for the ...
\end{enumerate}

The empirical record concerning facilitatory effects in reflexive agreement is mixed. 
Some have claimed that  such effects do not arise
\cite{sturt2003time,xiang2009illusory,dillon2013contrasting}, 
and that this is expected under a model in which the structural constraints from binding theory \cite{chomsky1982some} serve to effectively filter candidates for retrieval---in short, the parser does not consider or make contact with the ungrammatical distractor noun phrases \cite{sturt2003time,dillon2013contrasting}. 



\begin{figure}
    \centering
    \includegraphics[width=7cm]{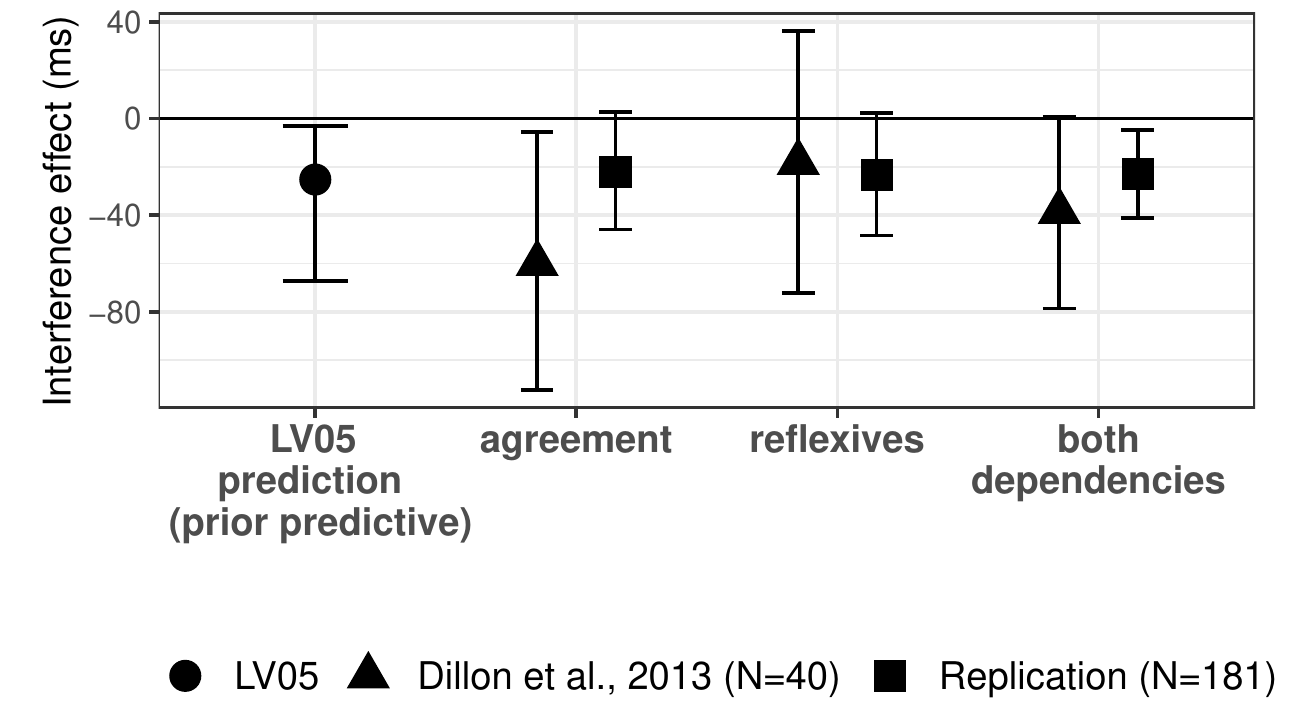}
    \caption{From \citet{jager2020interference}. Posterior estimates of facilitatory interference effects in subject-verb and reflexive agreement processing in a large scale replication of \citet{dillon2013contrasting}, the original effects, and predictions from the \citet{lewis2005activation} model.}
    \label{fig:jaeger-graph}
\end{figure}

However, a recent Bayesian meta-analysis of  key experiments by \citet{dillon2013contrasting} indicates substantially overlapping posterior estimates of facilitatory effects for subject-verb agreement and reflexive agreement \cite{vasishth0AD}. Concerned again about under-powered studies, \citet{jager2020interference} undertook a large scale (181 participants) eye-tracking replication and did find evidence for nearly equivalent facilitatory speed-ups for reflexive and subject-verb agreement (Figure~\ref{fig:jaeger-graph}). This result is not inconsistent with the meta-analysis, but provides stronger evidence that the facilitation effects in reflexives are real.

We take advantage of the very broad coverage of GPT-2 by having GPT-2 process the same set of sentence materials as human subjects in four different agreement experiments. To anticipate our key results, we find GPT-2 yields lower surprisal, i.e.\ facilitatory effects, in both subject-verb and reflexive pronoun conditions. 
Furthermore, we show that attention at the verb or pronoun is distributed to both target and distractor in just those conditions where the distractor matches the hypothesized number retrieval cue \cite{lin2019open}. Finally, we show that the surprisal contrasts between matching and nonmatching distractors in the grammatical (inhibitory) interference conditions are essentially zero.

\section{GPT-2 for psycholinguistic analysis}

\paragraph{The psycholinguistic relevance of GPT-2 and its training method.}
GPT-2 (Generative Pre-trained Transformer-2), introduced by OpenAI in \citet{radford2019language}, is a language model with a decoder-only Transformer architecture \cite{vaswani2017attention}, and 
has achieved state-of-the-art performance in diverse downstream tasks.
GPT-2 and other large-scaled language models based on transformer architectures were trained on billions of words of text, and engineered with performance in mind, not with concern for psycholinguistic plausibility. Why then should we then take them seriously as the basis of psycholinguistic models?

We believe that the new transformer-based models have three important properties that make them of psycholinguistic interest. (a) The models are  among the first to serve as the basis of systems that achieve human-level performance on a range of linguistic tasks, and they directly generate a key quantity, \emph{surprisal of the next word}, that we know is an important predictor of reading times in humans \cite{hale2001probabilistic,levy2008expectation}. (b) Although the data requirements are currently much greater than that for human language acquisition, the models are trained on a simple task---predict the next word---that may plausibly serve as the basis of a self-supervised learning signal in human language acquisition. The representations that arise from such learning are thus psycholinguistically interesting. (c) The  learned soft-attention and parallel content-based retrieval of   representations of prior input are architectural properties of the GPT models that align very closely with retrieval-based models of sentence comprehension \cite{lewis2006computational}. And the structure of these psycholinguistic models was proposed as a response to the challenges of computing long-distance dependencies---the same challenge that motivated the transformer as a departure from standard recurrent architectures \cite{vaswani2017attention, galassi2020attention}.

\begin{table*}[tb!]
\caption{A set of data included for the experiment on subject-verb agreement.
(\citet{wagers2009agreement}'s Exp3 also included sets with plural subjects in the ungrammatical conditions.)}
\label{tab:example_set}
\centering
\setlength\arrayrulewidth{1pt}

\begin{tabular}{cccl}
\hline
& \textbf{Interference} & \textbf{Grammaticality} & \textbf{Example sentences}\\
\hline
&int& gram & The \underline{commentator} who the \textbf{viewer} \textbf{trusts}  ... \\
\textbf{Wagers 2009}&non-int & gram & The \underline{commentators} who the \textbf{viewer} \textbf{trusts}  ...  \\
\textbf{Exp 2-3} &int & ungram & *The \underline{commentators} who the \textbf{viewer} \textbf{trust}  ... \\
&non-int & ungram & *The \underline{commentator} who the \textbf{viewer} \textbf{trust}  ...  \\
\hline
&int& gram & The \textbf{slogan} on the \underline{poster} \textbf{was} designed ... \\
\textbf{Wagers (2009)}&non-int & gram & The \textbf{slogan} on the \underline{posters} \textbf{was} designed ... \\
\textbf{Exp 4-6} &int & ungram & *The \textbf{slogan} on the \underline{posters} \textbf{were} designed ... \\
&non-int & ungram & *The \textbf{slogan} on the \underline{poster} \textbf{were} designed ... \\
\hline
&int& gram & The \textbf{executive} who oversaw the middle \underline{manager}  \\
&&& apparently \textbf{was} dishonest ...\\

& non-int & gram & The \textbf{executive} who oversaw the middle \underline{managers}  \\ \textbf{Dillon 2013}&&& apparently \textbf{was} dishonest ... \\

\textbf{Exp 1 agrmt} &int & ungram & *The \textbf{executive} who oversaw the middle \underline{managers}  \\ &&& apparently \textbf{were
} dishonest ... \\
&non-int & ungram &*The \textbf{executive} who oversaw the middle \underline{manager}  \\ &&& apparently \textbf{were} dishonest ... \\
\hline
\end{tabular}

\end{table*}

\paragraph{Identifying specialized heads in GPT-2.}
Here we use the medium-sized GPT-2 which is constructed with 12 layers, each of which includes 12 attention heads. Previous studies have revealed that 
individual attention heads in Transformer models serve are at least partially specialized in function \cite{clark2019does,vig2019multiscale,vig2019analyzing,voita2019analyzing}. 
Specifically, \citet{voita2019analyzing} found that certain attention heads are specialized for different dependency relations.

Following \citet{voita2019analyzing}'s method, 
we identified heads that are specialized for subject-verb relations and reflexive anaphora resolution.
\citet{voita2019analyzing}'s method works as follows.
First, sentences are parsed 
using CoreNLP dependency parser \cite{manning2014stanford}. Then, relative string positions (e.g., one token back, two tokens back) of all instances in each syntactic dependency were counted. Considering the proportion of the most frequent relative position as the baseline, attention heads are selected as specialized for a particular dependency relation if attention is paid for the corresponding dependent at least 10\% more often than the baseline. In other words, there must be some evidence that the attention head is sensitive to the dependency and not merely string position.


To find attention heads responsible for the relation between subjects and verbs, we used the CoreNLP parser on 148,376 sentences from the Brown corpus and Gutenberg corpus provided via Natural Language Toolkit (NLTK) \cite{bird2009natural}, extracting 49,145 \textit{nsubj} relations, which associate nominal subjects and their governors which are mostly verbs.
The most frequent relative position for \textit{nsubj} dependency relation is -1, which means that the nominal subjects usually come right before their governor, taking up 42\% of the cases. 

After analyzing the attention distribution pattern using GPT-2, we obtained four syntactic heads that were found to be partly specialized for \textit{nsubj} dependency relations: \textit{head4\_3} (59\%);  \textit{head3\_6} (51\%);  \textit{head6\_0} (49\%); \textit{head2\_9} (49\%)\footnote{head\textit{n\_m} refers to the \textit{m}-th attention head in the \textit{n}-th layer. Numbers in parentheses indicate accuracies of heads in paying the highest attention to the subject/antecedent by the verb/pronoun.}. Although we expect that the four syntactic heads responsible for \textit{nsubj} dependency relation may play distinct roles, in our analyses here we simply use the best performing head (\textit{head4\_3}).


The same method was implemented to find attention heads responsible for reflexive anaphora resolution. The only difference was that we used NeuralCoref \cite{wolf2018neuralcoref} to 
count relative position of antecedents to reflexive anaphora since the dependency parser does not associate antecedents and anaphora.
Out of 2,660 sentences that includes reflexive anaphora, we extracted 510 sentences where  NeuralCoref identified a single unique antecedent for the reflexive pronoun.
The most frequent relative position for reflexive anaphora and their antecedents was -2, meaning that antecedents appear before reflexive anaphora having one word in between. The proportion of the highest relative position was 22\%, requiring 24.2 \% of accuracy for attention heads to be considered responsible for reflexive anaphora resolution. We found four heads whose accuracies are higher than the threshold: \textit{head1\_5} (44\%);  \textit{head3\_5} (39\%);  \textit{head4\_3} (27\%); \textit{head6\_0} (25\%), and we again take the best performing head (\textit{head1\_5}) for further analysis. 

\paragraph{Metrics.}

\label{sec:metrics}
We define here three metrics for our analyses:
\textit{surprisal}, \textit{attention entropy from syntactic heads}, and \textit{attention to target}. 
We use surprisal for making reading time predictions, but use the attention metrics to provide insight into the processing at the critical region and therefore the representations computed in the prefix  before the critical region.
{Surprisal} is thus based on the final prediction of the entire model, but the attention metrics are associated with the attention heads most specialized for our dependencies of interest.

\textbf{Surprisal} \cite{hale2001probabilistic, levy2008expectation} is defined as the negative log probability of the word given left context. 
\begin{equation}
    \mathrm{Surprisal}(w) = -{\log_2}P(w|context)
\end{equation}
Any use of surprisal requires adoption of some kind of language model; e.g. some past work has used probabilistic CFGs \cite{levy2008expectation}. Here we use GPT-2, which computes after each word a probability distribution over its large lexicon that is conditioned on its internal representation of the left context.


\textbf{Attention to target} is simply the value of the soft attention vector element that corresponds to the target word position, which we denote $\mathrm{Attn}({w_{cue}},{w_{target}})$, and indicates how much attention is allocated to the target by one of the specialized attention heads (\textit{head4\_3} for subject-verb and \textit{head1\_5} for reflexives.)

\textbf{Attention entropy} is a variant of \citet{shannon1948mathematical}'s information entropy that we use as a measure of how sharply focused (low entropy) or diffuse (high entropy) the attention pattern is. (It may be thought of as a measure of the uncertainty about the attentional target, but because the attention values are not probabilities from which targets are sampled, this interpretation is not strictly warranted).

\begin{fleqn}[\parindent]
\begin{equation}
\begin{aligned}
\mathrm{Entropy}({w_i}) = \hspace{120pt}\\
    \sum_{j=1}^{i-1}\mathrm{Attn}({w_i},{w_j}) \times{\log_2}\mathrm{Attn}({w_i},{w_j})
\end{aligned}
\end{equation}
\end{fleqn}
where \textit{i} refers to the location of the critical word, \textit{j} are  locations of prior words, and $\textrm{Attn}(w_{i}, w_{j})$ is attention allocated to  $w_j$ from $w_i$.




\section{Subject-verb agreement experiments}

To investigate whether GPT-2 may predict facilitatory interference effects in subject-verb agreement, we  ran GPT-2 on materials from three studies
\cite{dillon2013contrasting, wagers2009agreement}: 48 sets of sentences from Experiments 2-3 in \citet{wagers2009agreement}\footnote{\citet{wagers2009agreement}'s materials are an extended and slightly modified version of \citet{pearlmutter1999agreement}}; 24 sets of sentences from Experiments 4-7 in \citet{wagers2009agreement}; 48 sets of sentences from \citet{dillon2013contrasting} (See Table~\ref{tab:example_set}). 
\par
These three sets of sentences have in common a $2\times2$ structure with the factors  \textit{grammaticality} (grammatical/ungrammatical) and \textit{interference} (interfering/non-interfering), as described above.
Additionally, \citet{wagers2009agreement}'s Exp 3 also includes an additional condition, \textit{subject} (singular/plural) for  investigating a possible singular-plural asymmetry, i.e., asking whether interference effects are equivalent for plural (for plural verbs) and singular (for singular verbs) distractors. 

Note that sentences from Experiments 2--3 in \citet{wagers2009agreement} involve structures in which the distractor appears \emph{before} the target, and so test effects of \emph{proactive interference}. Thus the distractors are also more distant from verbs than in the other experimental materials.

\begin{figure}[tb!]
\centering
\begin{subfigure}[b]{8cm}
\begin{center}
\includegraphics[width=6.4cm]{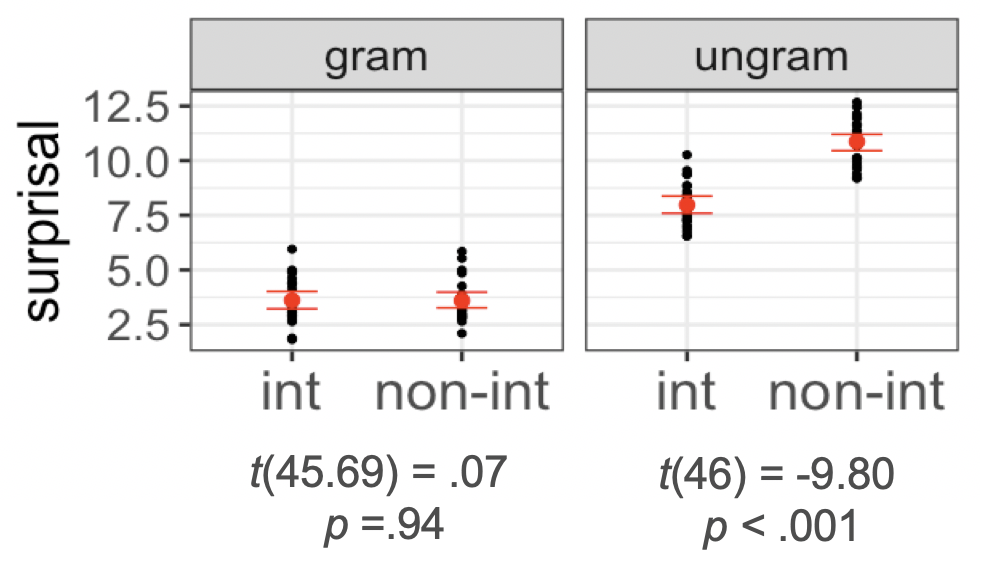}
\caption{Wagers et al. 2009 (Exp 4--6).}
\label{fig:subj_verb3}
\end{center}
\end{subfigure}



\begin{subfigure}[b]{8cm}
\begin{center}
\includegraphics[width=6cm]{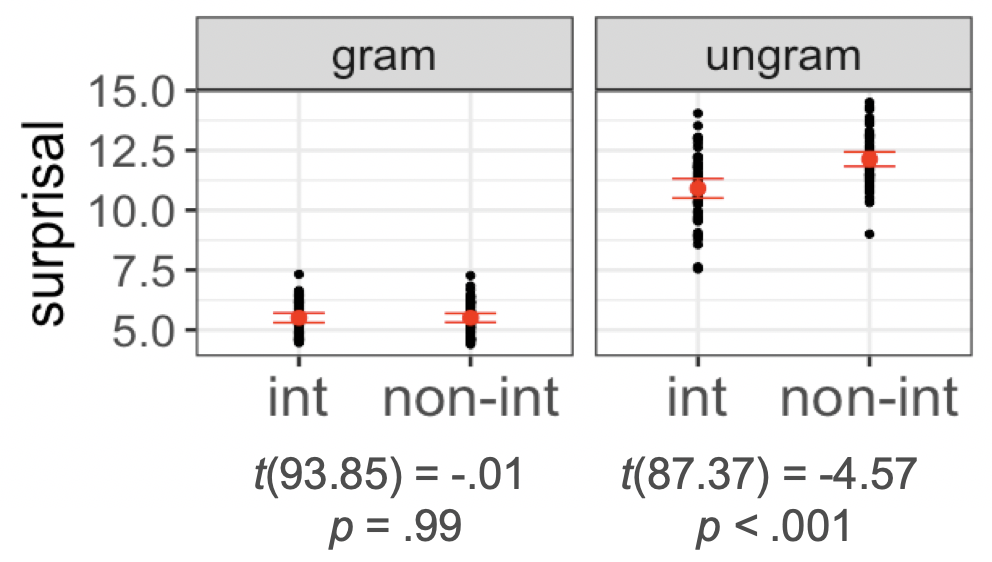}
\caption{Dillon et al. 2013 (Exp 1)}
\label{fig:subj_verb4}
\end{center}
\end{subfigure}

\medskip

\begin{subfigure}[b]{8cm}
\begin{center}
\includegraphics[width=6cm]{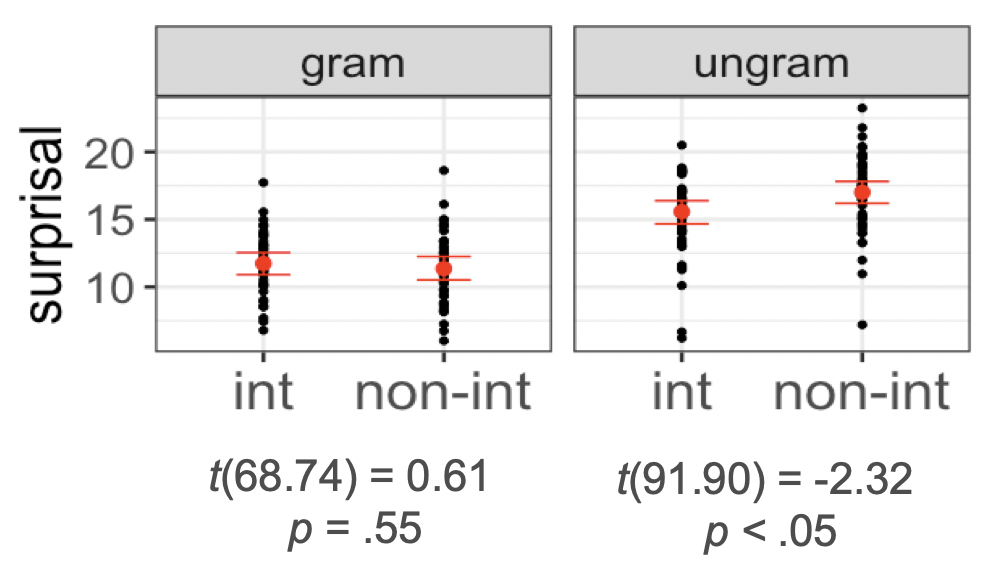}
\caption{Wagers et al. 2009 (Exp 2--3, singular subject)}
\label{fig:subj_verb2}
\end{center}
\end{subfigure}

\medskip

\begin{subfigure}[b]{8cm}
\begin{center}
\includegraphics[width=6cm]{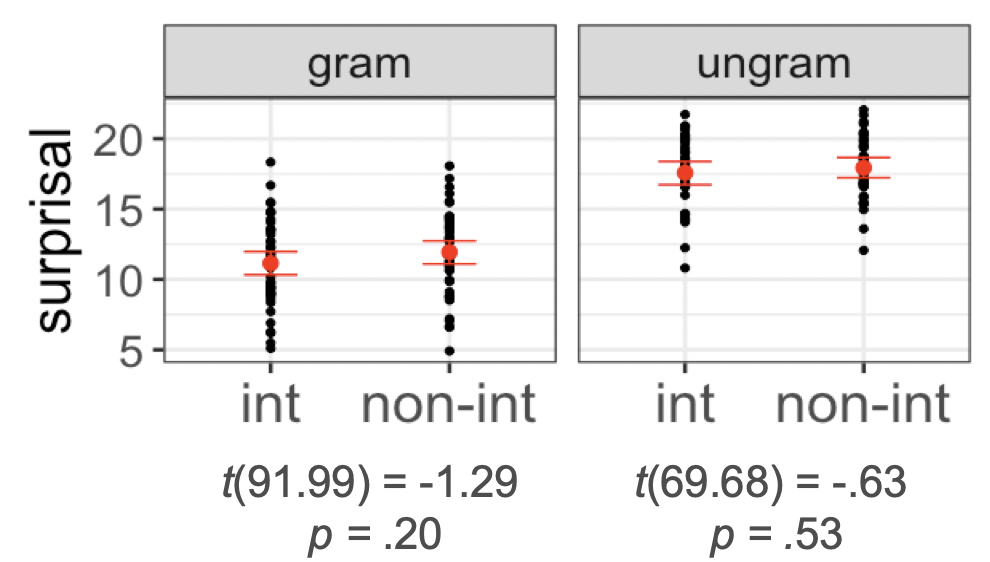}
\caption{Wagers et al. 2009 (Exp 3, plural subject)}
\label{fig:subj_verb1}
\end{center}
\end{subfigure}

\caption{The surprisal of critical verbs computed by GPT-2 on the materials in four subject-verb number agreement experiments. Each small dot is a data point from one sentence; the red dots and intervals represent means and 95\% confidence intervals.}
\label{fig:subject-verb-surprisal}
\end{figure}

\paragraph{Results of surprisal analyses.}
Figure \ref{fig:subject-verb-surprisal} shows the surprisal computed at the critical verbs in each of the experiments and in each of the four conditions separately (red dots and intervals represent means and conventional 95\% confidence intervals).
Surprisal matches the important qualitative pattern found in the meta-analysis of first-pass reading times:  lower surprisal---facilitatory effects---are found in the ungrammatical conditions when the distractor matches the verb's number, and no inhibitory effects are found in the grammatical conditions. Furthermore, the effects are largest for the case of \emph{retroactive} interference, where the distractor follows the target and immediately precedes the verb (Figure~\ref{fig:subj_verb3}), compared to \emph{proactive} inteference, where the distractor precedes the target (Figure~\ref{fig:subj_verb2}). 
The exception is that no facilitatory effects were found when the verb is singular and the target subject is plural
(see Figure~\ref{fig:subj_verb1}). But the facilitatory effect in this condition was not reliably different from zero in the meta-analysis, and it mirrors a  plural-singular asymmetry (or \emph{markedness} effect) found in agreement attraction in production.
\par


\begin{figure*}[tbh!]
\centering
\begin{tabular}{|>{\centering\arraybackslash}m{3cm}| >{\centering\arraybackslash}m{5cm}|>{\centering\arraybackslash}m{5cm}|}
\hline
 & \centered{\textbf{interfering}} & \textbf{non-interfering}\\
\hline
\textbf{grammatical} & \includegraphics[scale = 0.39]{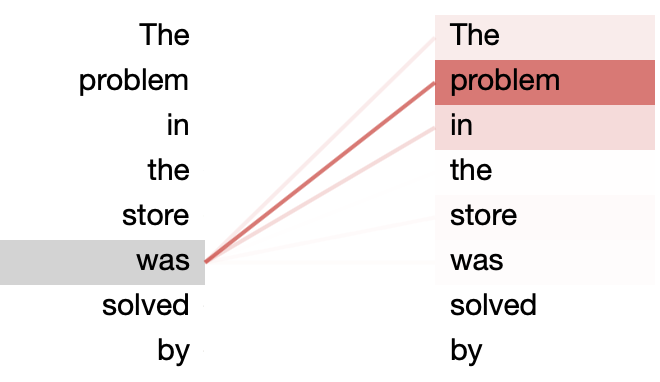} & \includegraphics[scale = 0.4]{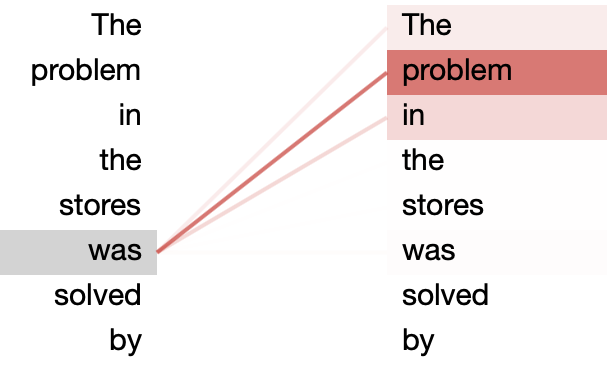} \\
\hline
\textbf{ungrammatical} & \includegraphics[scale = 0.39]{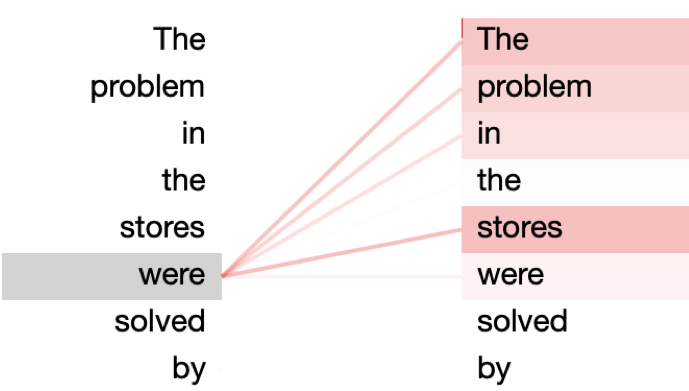} & \includegraphics[scale = 0.4]{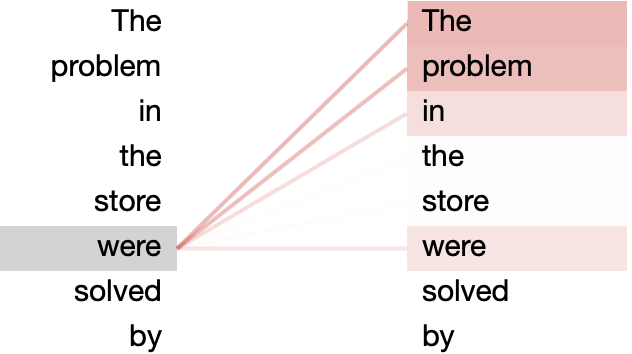}\\
\hline
\end{tabular}
\caption{\label{subject_verb_attn}
An example of the attention distribution of an attention head specialized for subject-verb dependencies in the four conditions of the subject-verb agreement experiments.}
\label{fig:attn_distribution}
\end{figure*}

\begin{figure*}[h!]
\centering

\begin{center}
\includegraphics[width=6cm]{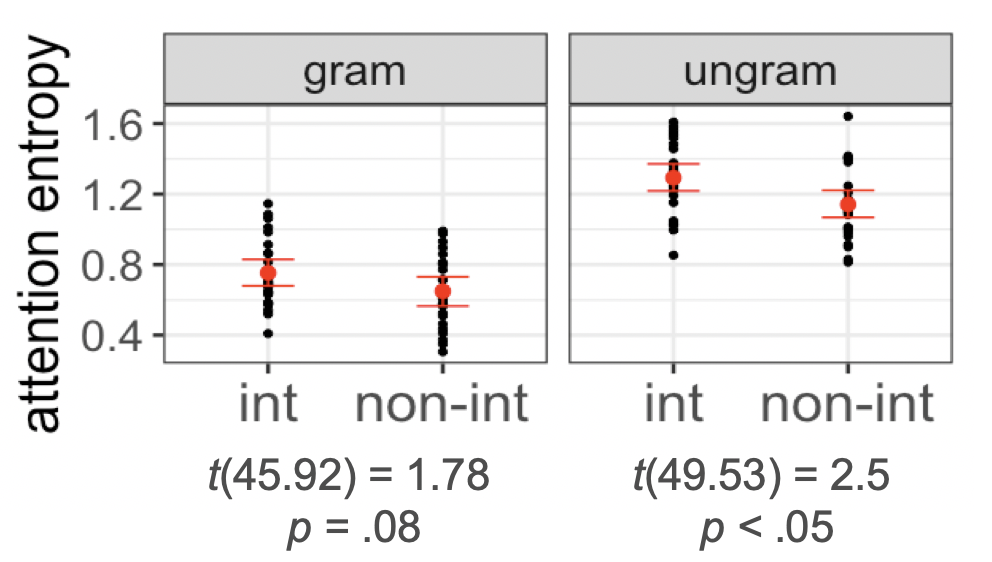}
\hspace{0.2cm}
\includegraphics[width=6cm]{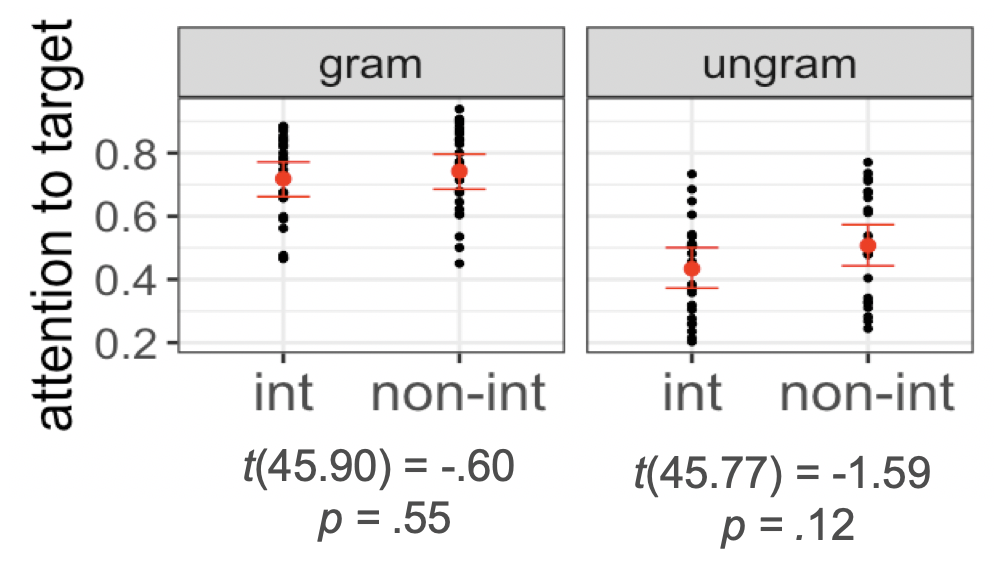}
\newline
(a) Wagers et al. 2009 (Exp 4--6)
\end{center}
\vspace*{0.1cm}

\begin{center}
\includegraphics[width=6cm]{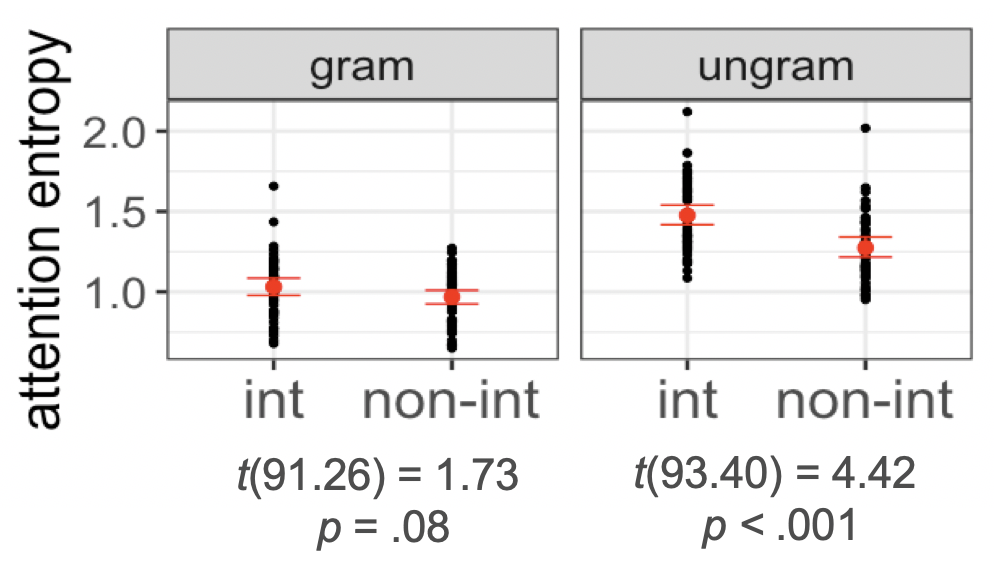}
\hspace{0.2cm}
\includegraphics[width=6cm]{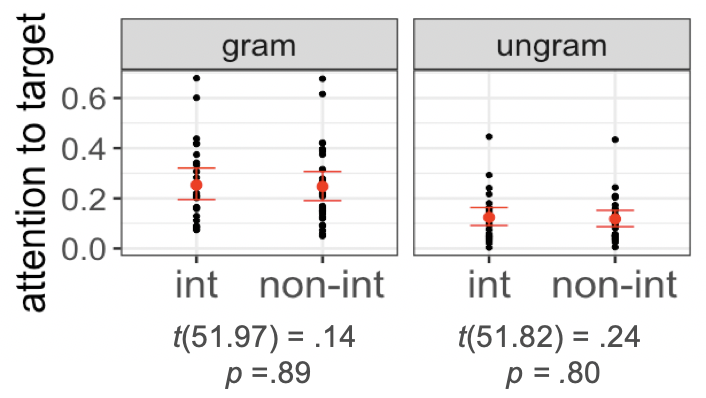}
\newline
(b) Dillon et al. 2013 (Exp 1)
\end{center}

\vspace*{0.1cm}

\begin{center}
\includegraphics[width=6cm]{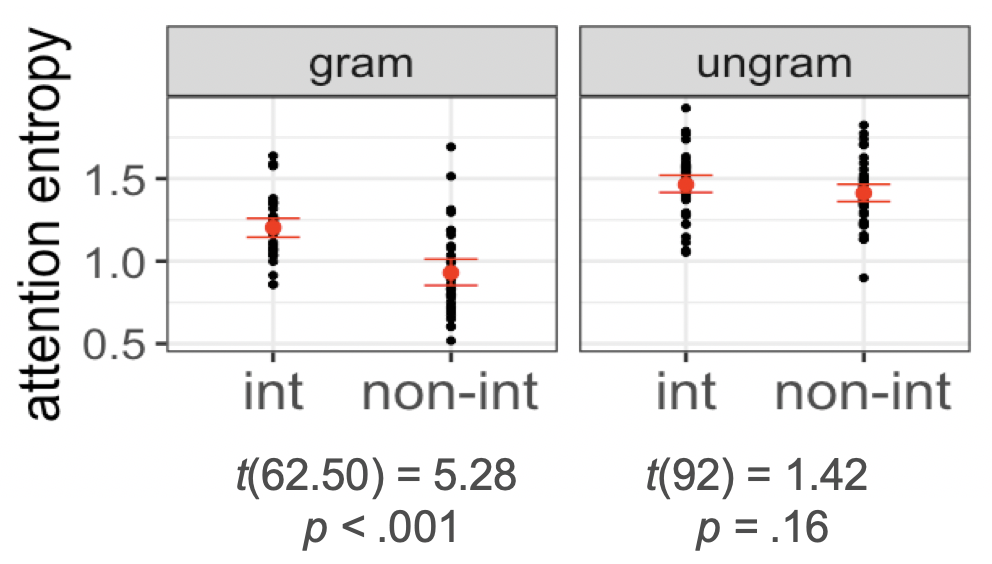}
\hspace{0.2cm}
\includegraphics[width=6cm]{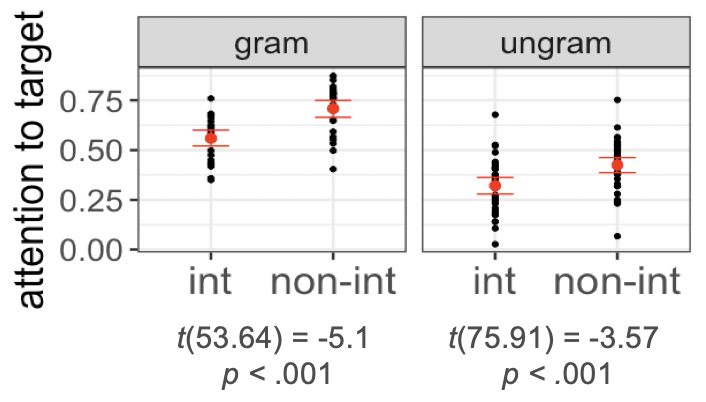}
\newline
(c) Wagers et al. 2009 (Exp 3, singular subject)
\end{center}

\vspace*{0.1cm}

\begin{center}
\includegraphics[width=6cm]{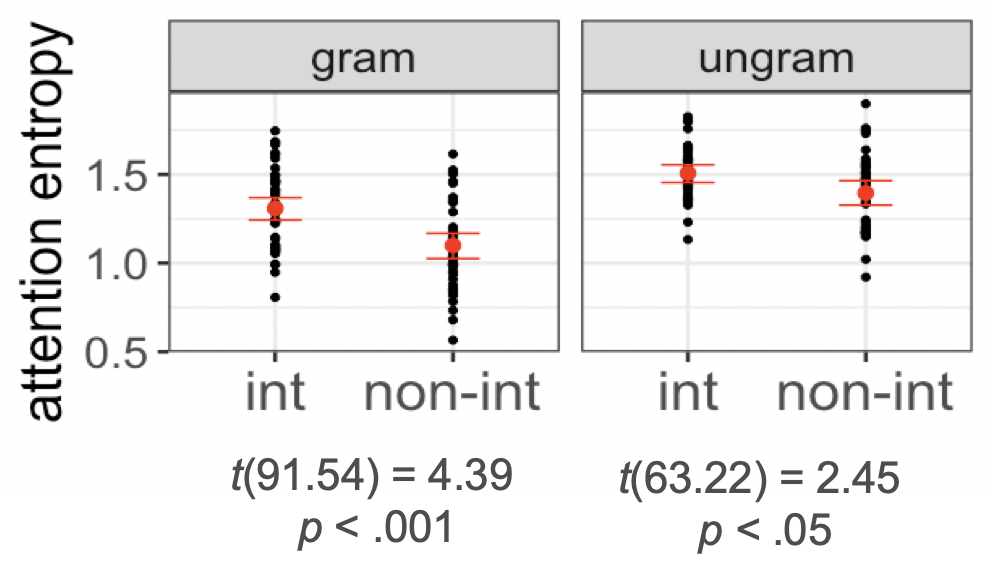}
\hspace{0.2cm}
\includegraphics[width=6cm]{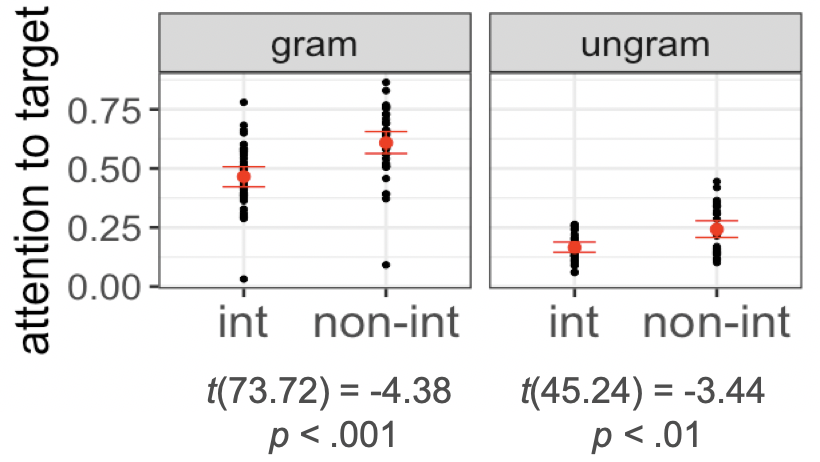}
\newline
(d) Wagers et al. 2009 (Exp 3, plural subject)
\end{center}

\caption{Metrics quantifying attention patterns of the attention head most specialized for subject-verb relations, computed at the verb in the subject-verb agreement experiments.}
\label{fig:subj-verb-attn}
\end{figure*}

\paragraph{Results of attention analyses.}
Our conjecture is that in the \emph{interfering} conditions where the distractor matches the verb in number
 that the attention of the \emph{nsubj}-specialized attention head \emph{head4\_3} will be distributed to both the target \emph{and} the distractor. It is possible to visualize exactly this pattern using a tool developed by 
 \citet{vig2019multiscale}. Figure~\ref{fig:attn_distribution} shows an example visualization.
 
Analyses of the \textit{attention entropy} and \textit{attention to target} metrics  provide quantitative evidence for this conjecture: 
Figure~\ref{fig:subj-verb-attn} shows two metrics across the four datasets. 
The interfering conditions always show the highest value of \textit{attention entropy} and the lowest value of \textit{attention to target}, which means that the head most specialized for subject-verb relations 
distributes attention more
diffusely and away from the target subject. There is evidence for the expected attention effects even in the grammatical conditions, but in these conditions there is no effect of surprisal.  
Thus, under a theory in which similarity-based interference exerts its effects on reading time through a \emph{surprisal bottleneck} \cite{levy2008expectation}, no reading time differences are expected here---even though the underlying representations and attention patterns may reflect the interference.



\paragraph{Preliminary corpus analysis of ungrammatical subject-verb agreement sentences.}
One possible explanation for the observed facilitatory interference effects is that GPT-2 was exposed to 
ungrammatical sentences in the training data that have precisely the interference patterns of the ungrammatical sentences in our experiments. To examine such possibility, we analyzed 241 sentences randomly extracted from a Reddit corpus \cite{chang2020convokit} whose subjects and verbs do not agree in number, and have either interfering or non-interfering distractors in between. The results shown in Table \ref{tab:corpus} suggest that interfering distractors occur about twice as often as non-interfering distractors in the case of singular subjects with an ungrammatical plural verb, consistent with our expectations that agreement-attraction errors in production may be evident in un-edited corpora. 

But it seems unlikely that this 2:1 ratio, which corresponds to about a 1 bit difference in surprisal, is sufficient alone to explain the observed surprisal differences. For example, in the Wagers et al Experiment 4--6, we observed about a 3 bit difference in surprisal, a 2 bit or 4x difference in probability  relative to what would be expected on the basis of the corpus counts. More extensive corpus analysis is necessary to confidently rule out this explanation.

\begin{table}
\begin{center}
\begin{tabular}{ccc} 
\hline
 & {singular subj} & {plural subj}\\
\hline
{interfering} & 80 & 71  \\
{non-interfering} & 39 & 51 \\
\hline
\end{tabular}
\end{center}
\caption{Results from a preliminary corpus analysis of patterns of ungrammatical subject-verb agreement. In the key case of a singular subject and a plural verb, the number of an intervening distractor is about twice as likely to be plural (interfering) rather than singular (non-interfering). See text for a discussion.}
\label{tab:corpus}
\end{table}

\section{Reflexive agreement experiments}

To examine whether the prediction of GPT-2 are consistent with the null interference effects argued for by \citet{dillon2013contrasting}, or show facilitatory interference effects as in the large scale \citet{jager2020interference} replication, we conducted an experiment using the same methodology as described above for the subject-verb experiments, but using the reflexive materials in  \citet{dillon2013contrasting}, and focusing the attention analyses on the head most specialized for reflexive anaphor resolution. Examples of the materials are shown in Table \ref{tab:example_set_ref}.

\begin{figure}[t!]
\centering
\includegraphics[width=6.1cm]{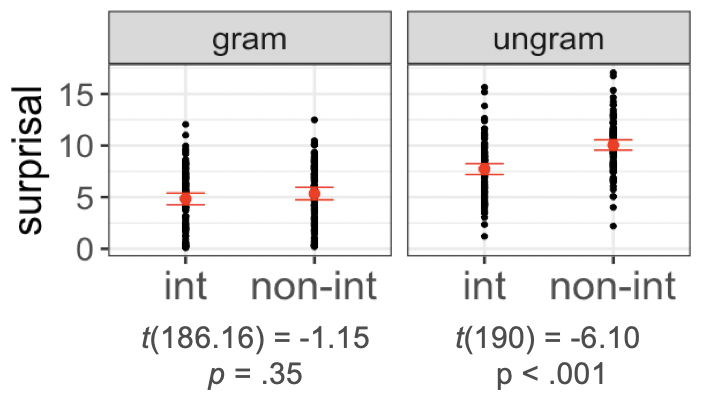} 
\includegraphics[width=6.1cm]{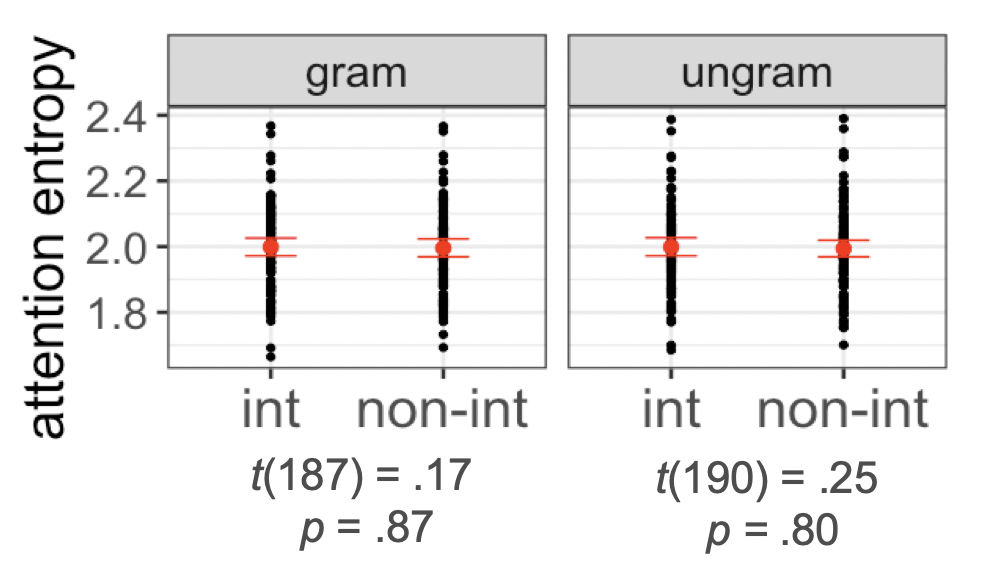}
\includegraphics[width=6.1cm]{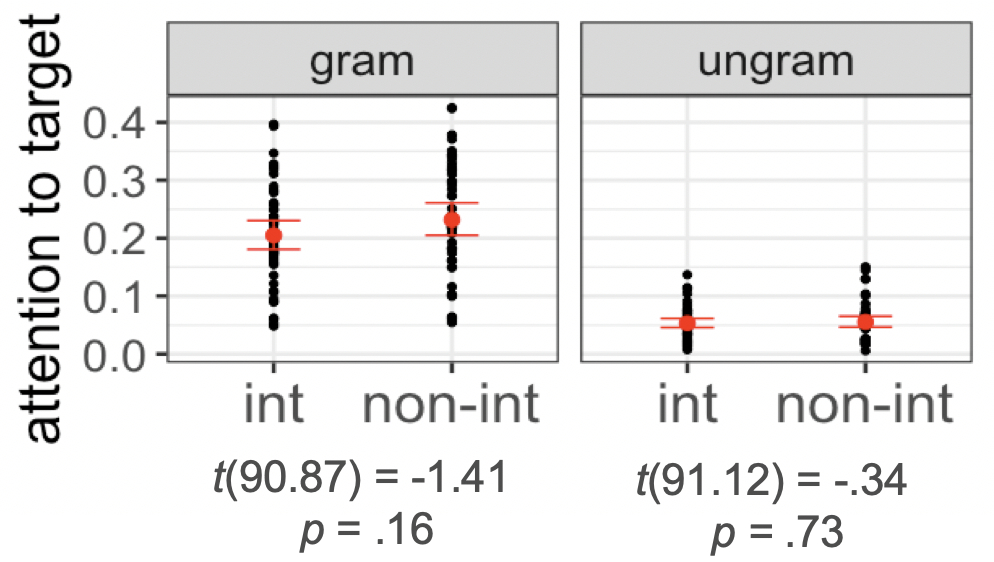}
\label{fig:reflexive_metrics}
\caption{Results of the GPT-2 reflexive agreement experiment using materials from \citet{dillon2013contrasting}.}
\label{fig:reflexive}
\end{figure}

\paragraph{Results of the surprisal analyses.}
Summaries of the surprisal (and attention metrics)  measured at reflexive anaphora are provided in Figure \ref{fig:reflexive}.
Consistent with the large scale replication of \citet{dillon2013contrasting} conducted by \citet{jager2020interference} (but inconsistent with the null results reported by Dillon et al), we found lower \textit{surprisal} values in the ungrammatical interfering conditions, consistent with a facilitatory interference effect. 
\begin{table*}[tb!]
\centering
\setlength\arrayrulewidth{1pt}
\begin{tabular}{cccl}
\hline
& \textbf{Interference} & \textbf{Grammaticality} & \textbf{Example sentences}\\

\hline
&int& gram & The basketball \textbf{coach} who trained the star \underline{player} \\
&&& usually blamed \textbf{himself} for the ...\\

& non-int & gram & The basketball \textbf{coach} who trained the star \underline{players} \\ \textbf{Dillon 2013}&&& usually blamed \textbf{himself} for the ... \\

\textbf{Exp 1 reflexive} &int & ungram & *The basketball \textbf{coach} who trained the star \underline{players} \\ &&& usually blamed \textbf{themselves} for the ... \\
&non-int & ungram &*The basketball \textbf{coach} who trained the star \underline{player}  \\ &&& usually blamed \textbf{themselves} for the ... \\
\hline
\end{tabular}
\caption{Examples from \citet{dillon2013contrasting}, used in the GPT-2 experiment on reflexive pronoun agreement.}
\label{tab:example_set_ref}
\end{table*}

\paragraph{Results of the attention analyses.}
We found little or no differences between interfering and non-interfering cases in the  two attention metrics \textit{attention entropy} and \textit{attention to target}. It is possible that this is because the attention head \textit{head1\_5} that we found to be partly specialized for reflexive anaphora resolution is actually not as specialized in reflexive anaphora resolution as \textit{head4\_3} specialized in \textit{nsubj} dependency resolution. We cannot conclude yet whether there exist heads that serve this function better (that are not detected by the method of \citet{voita2019analyzing}), whether GPT-2 is not reliably resolving the reflexive anaphora, or whether GPT-2 is doing so in a way that is distributed across many attention heads.


\section{Discussion and future directions}

Effects of similarity-based interference have been the province of models of noisy memory rather than models of probabilistic expectations, because in standard probabilistic grammars the expectation for the agreement features of a licensor such as a verb or pronoun should not be conditioned upon the agreement features of constituents other than the target licensee. But we show here that 
a large-scale Transformer language model, GPT-2, trained only to predict the next word, nevertheless yields surprisal values that 
are consistent with facilitatory interference effects due to distractor noun phrases that do not participate in the agreement relations.
We also confirmed that two metrics that are easily computed from the Transformers' attention mechanism, \textit{attention entropy} and \textit{attention to target}, show patterns in the subject-verb experiments that are consistent with cue-based retrieval models. 

Our results are suggestive of a possible interesting link between surprisal and noisy memory representations. The attention patterns that we have discovered must reflect  similarity between the representations of the target and distractor noun phrases. This representational similarity is the source of great generalization power,  but this generalization can lead to linguistic expectations that are not derived by conventional grammatical analyses.
\par

One limitation of our analyses of attention is that they depend on methods for identifying specialized heads for specific dependency types. It is not clear that we understand enough about Transformer models to do this reliably. 
But our results suggest that for at least some dependencies, these simple attention metrics and head selection methods can yield interesting insights.
\par
The approach outlined may provide an important way to combine surprisal and noisy memory accounts, maintaining a surprisal bottleneck. Using trained Transformers has the significant theoretical advantage that the memory representations, the attention/retrieval cues, and thus the predicted similarity effects are \emph{learned} via a  self-supervised prediction task. And so such models naturally yield experience-driven sources of noisy representations that are independent of the process noise assumed in existing memory-based models. Combining the process- and experience-based noise in a single model is an important goal for psycholinguistic theory.

\bibliography{naacl2021}
\bibliographystyle{acl_natbib}



\end{document}